 \documentclass[preprint, 10pt, a4paper]{elsarticle}

\usepackage{amssymb}

\usepackage{float}
\restylefloat{table}

\usepackage[utf8]{inputenc}
\usepackage[T1]{fontenc}
\usepackage{textcomp}
\usepackage{soul}
\usepackage{graphicx,subfig} 
\usepackage{amsmath,array,amssymb,xparse,upgreek,bm} 
\usepackage{tikz,mathpazo} 
\usetikzlibrary{shapes.geometric,arrows} 
\usepackage{hyperref}
\usepackage{tcolorbox} 
\usepackage{booktabs}
\usepackage{multirow}

\usepackage[ruled]{algorithm2e}
\usepackage{empheq}
\usepackage{tablefootnote}
\usepackage{array}

\journal{SoftwareX}

\begin{document}

\begin{frontmatter}




\title{SerialTrack: ScalE and Rotation Invariant Augmented Lagrangian Particle Tracking}

\author{Jin Yang (0000-0002-5967-980X) \fnref{1}}
\author{Yue Yin \fnref{1,2}}
\author{Alexander K. Landauer (0000-0003-2863-039X) \fnref{3}}
\author{Selda Buyuktozturk \fnref{1,4}}
\author{Jing Zhang \fnref{1}}
\author{Luke Summey \fnref{1}}
\author{Alexander McGhee \fnref{1}}
\author{Matt K. Fu \fnref{5}}
\author{John O. Dabiri \fnref{5}}
\author{Christian Franck \corref{cor1}\fnref{1}}
\cortext[cor1]{Corresponding author}
\ead{cfranck@wisc.edu}
\address[1]{University of Wisconsin-Madison, Mechanical Engineering, Madison, WI, USA}
\address[2]{Department of Mechanical Engineering, Carnegie Mellon University, Pittsburgh, PA, USA}
\address[3]{National Institute of Standards and Technology, Gaithersburg, MD, USA}
\address[4]{School of Engineering, Brown University, Providence, RI, USA}
\address[5]{Graduate Aerospace Laboratories, California Institute of Technology, Pasadena, CA, USA}


\begin{abstract}
We present a new particle tracking algorithm to accurately resolve large deformation and rotational motion fields,  which takes advantage of both local and global particle tracking algorithms. We call this method the ScalE and Rotation Invariant Augmented Lagrangian Particle Tracking (SerialTrack). 
This method builds an iterative scale and rotation invariant topology-based feature for each particle within a multi-scale tracking algorithm. The global kinematic compatibility condition is applied as a global augmented Lagrangian constraint to enhance the tracking accuracy. An open source software package implementing this numerical approach to track both 2D and 3D, incremental and cumulative deformation fields is provided.
\end{abstract}


\end{frontmatter}

\section*{Required metadata}
\label{}

\section*{Current code version}
\label{}

\begin{table}[H]
\begin{tabular}{|l|>{\raggedright\arraybackslash}p{5cm}|>{\raggedright\arraybackslash}p{6.3cm}|}
\hline
\textbf{Nr.} & \textbf{Code metadata description} & \textbf{Please fill in this column} \\
\hline
C1 & Current code version & v1.0 \\
\hline
C2 & Permanent link to code/repository used for this code version & https://github.com/FranckLab/   SerialTrack  \\
\hline
C3 & Code Ocean compute capsule &  \\ 
\hline
C4 & Legal Code License   & MIT license  \\
\hline
C5 & Code versioning system used & git \\ 
\hline
C6 & Software code languages, tools, and services used &  MATLAB\footnotemark \\ 
\hline
C7 & Compilation requirements, operating environments \& dependencies &  
MATLAB with the following toolboxes: System Identification Toolbox,
Image Processing Toolbox,
Statistics and Machine Learning Toolbox,
Partial Differential Equation Toolbox,
Wavelet Toolbox,
Curve Fitting Toolbox,
Parallel Computing Toolbox,
MATLAB Parallel Server, and
Polyspace Bug Finder\\
\hline
C8 & If available Link to developer documentation/manual &  https://github.com/FranckLab/   SerialTrack/wiki \\
\hline
C9 & Support email for questions & cfranck@wisc.edu  \\
\hline
\end{tabular}
\caption{Code metadata\\
\\
$^1$ Certain commercial equipment, software and/or materials are identified in this paper in order to adequately specify the experimental procedure. In no case does such identification imply recommendation or endorsement by the National Institute of Standards and Technology, nor does it imply that the equipment and/or materials used are necessarily the best available for the purpose.} 

\label{} 
\end{table}


%
%
%
%

\section{Motivation and significance}
\label{sec:intro}
%
%
%
%

Single particle tracking (SPT) and particle tracking velocimetry (PTV) methods provide quantitative, temporally resolved measurements of motions and deformations to investigate complex dynamics at the resolution of single tracking features by automatically localizing and tracking individual particles \cite{chenouard2014natmeth,maas_particle_1993,meijering_chapter_2012}. This is in contrast to digital image/volume correlation (DIC/DVC) and particle image velocimetry (PIV), which are subset-correlation-based techniques (see a summary of various full-field tracking methods in Table\,\ref{tab:sum_mesh}). The increased specificity of the tracked particles can be beneficial to applications such as quantitative biological motion tracking \cite{hazlett2020epifluorescence,ulman2017natmeth,ewers_single-particle_2005,sbalzarini_feature_2005,wu_collective_2006,leggett_mechanophenotyping_2020} and fluid mechanics for flow measurements \cite{ardekani_three-dimensional_2013,maas_particle_1993,kim_three-dimensional_2016,saha_breakup_2016}. However, it remains a challenge to uniquely and robustly match particles throughout an image sequence.

Particle tracking methods have been used to study multiple length and time scales in soft materials and rheology  \cite{crocker1996methods,mason_optical_1995}, experimental physics  \cite{manzo2015review}, materials science  \cite{huang2013imaging,novara_particle-tracking_2013}, and geophysics \cite{smith_submersible_2008}.  There are a number of SPT and PTV algorithms created for various applications \cite{pereira_two-frame_2006,ouellette_quantitative_2006} but they are often specialized, and typically require either small inter-frame deformations \cite{boltyanskiy_tracking_2017,legant_measurement_2010,schanz_shake--box_2016} or sparse \cite{cierpka_higher_2013,feng_adaptive_2014,jaqaman_robust_2008} /dense \cite{patel2018rapid,ohmi_particle_2009} particle seeding volumes to perform well.

Particle tracking procedures can generally be divided into two steps:  (i) \textit{particle localization}, where coordinates of individual particles are extracted from each frame of an image sequence, and (ii) \textit{particle linking}, where detected particles are uniquely matched from frame to frame to construct a motion field.

Particle localization algorithms often decompose the process into particle detection followed by subpixel centroid localization. For example, images are pre-processed to reduce noise and selectively enhance objects, then particle spots or feature locations are detected by applying image segmentation, local-maxima finding, or other thresholding criteria. Particle centroid locations are often estimated by applying Gaussian fitting \cite{abraham_quantitative_2009,small_fluorophore_2014} or intensity-based centroid measurements \cite{liu_fast_2013,parthasarathy_rapid_2012}. In general, all of the above image processing-based methods perform well for images with sufficiently high signal-to-noise ratios (SNR $\leq$ 5). Recently, machine learning-based methods have been developed that can potentially improve performance in images with spatiotemporal heterogeneity and poor signal-to-noise ratios \cite{newby2018convolutional,Ershov2021trackmate,zepeda_o_untying_2021}.

Various algorithms have been created to detect and track individual particles \cite{chenouard2014natmeth} such as the straightforward \textit{k}-nearest-neighbor (\textit{k}NN) searches, topology-based approaches where neighboring particles are employed to construct local surrounding topology features \cite{patel2018rapid,cui_three-dimensional_2018,rubbert_iterative_2020,zhang_particle_2015,heyman_tractrac_2019}, globally optimized search problems -- including linear assignment programming \cite{jaqaman_robust_2008}, Kalman filtering \cite{TINEVEZ2017trackmate}, relaxation methods \cite{pereira_two-frame_2006,ohmi_particle-tracking_2000}, and feature vector-based techniques \cite{legant_measurement_2010,feng_adaptive_2014} (see a brief summary of particle tracking open-source codes in Table\,\ref{tab:ptv_codes}).  Among these methods, the nearest neighbor-type search algorithms are typically suitable for relatively low numbers of particles that undergo displacements smaller than the typical interparticle separation distance. The more robust topology-based and feature-based particle tracking algorithms are able to resolve large deformation fields but favor large particle numbers. The relaxation-based approaches perform well on highly stochastic motion fields but require small inter-frame motions. The nearest-neighbor and local topology-based methods are computationally efficient and can be easily implemented in parallel. However, they both have limitations in regard to particle seeding densities and there is no guarantee that the final tracked motion fields are unique and kinematically admissible. Global optimization particle tracking methods can guarantee the uniqueness and kinematic admissibility of the tracked motion, but are typically computationally expensive.

Here we present a new particle tracking algorithm, called the \textbf{S}cal\textbf{E} and \textbf{R}otation \textbf{I}nvariant \textbf{A}ugmented \textbf{L}agrangian Particle Tracking (SerialTrack) method, which takes advantage of both local (i.e., nearest-neighbor search and local-topology-based feature tracking \cite{patel2018rapid,janke_part2track_2020}) and global optimization to reconstruct motion fields in either 2D or 3D, and with large, complex deformations for both sparse and dense particles  efficiently, robustly and accurately. This new method first builds a local scale and rotation invariant topology-based feature for each particle, then iteratively tracks these within a global multiscale algorithm. The global kinematic compatibility condition is applied as an augmented Lagrangian constraint \cite{yang2019augmented,yang2020augmented} to enhance the tracking accuracy. In addition to tracking incremental deformation between two subsequent frames, SerialTrack can track large cumulative deformations where the initial guess of each tracking displacement field has been improved by a data-driven reduced order modeling method \cite{yang2021staq}. The new method includes both particle localization and particle linking processes, and also may optionally account for shape distortion of particles due to large deformations.

\begin{table}[t!]
\centering
\caption{Comparison of different full-field measurement methods}
\label{tab:sum_mesh}
\resizebox{1\textwidth}{!}{
\begin{tabular}{>{\raggedright\arraybackslash}p{4.cm}p{1.5cm}p{1cm}p{1.cm}p{1.5cm}p{1.5cm}p{2.3cm}}
\toprule
Technique name  &  PIV & DIC & DVC & PTV & SPT & \textit{\textbf{SerialTrack (this work)}}  \\ \midrule
Considering particle shape distortion or not & No & Yes & Yes & No & No & Yes  \\[.75cm]
Matching continuous patches (C) or discrete feature points (D)  & C & D & C & C & D & D  \\[1.25cm]
2D image sequence or 3D volumetric image & 2D \& 3D & 2D & 3D & 2D \& 3D & 2D \& 3D & 2D \& 3D \\[.75cm]
Tracking velocity field (v) or cumulative displacement (u) & v & u & u & v & u & v \& u \\
\bottomrule 
\end{tabular}}
\end{table}

\begin{table}[t!]
\centering
\caption{Comparison of different open-source particle tracking codes}
\label{tab:ptv_codes}
\resizebox{1\textwidth}{!}{
\begin{tabular}{p{2.2cm}p{1.4cm}p{2.0cm}p{1.5cm}>{\raggedright\arraybackslash}p{7cm}}
\toprule
Name & Refs & Language &  Dim. & Particle linking algorithm \\ \midrule
PTVlab & \cite{brevis2011integrating,link_ptvlab}  &   MATLAB & 2D & Integrated cross-correlation and relaxation algorithm \\[.75cm]
OpenPTV & \cite{link_openptv} &  C \& Python & 3D &  A spatio-temporal matching algorithm \cite{willneff2003spatio} \\[.75cm]
TrackMate & \cite{Ershov2021trackmate,link_trackmate} &  MATLAB \quad \& Fiji & 2D \& 3D & LAP, u-track, Kalman filter, etc \cite{trackmate_alg_link}. \\[.75cm] %
TracTrac & \cite{heyman_tractrac_2019,link_TracTrac}  &  MATLAB \quad \&  Python & 2D & K-dimensional trees to search and compute statistics around
neighboring objects \\[.75cm]
TPT & \cite{patel2018rapid,link_tpt}  & MATLAB & 3D & Topology-based matching and iterative deformation warping (IDM) \\[.75cm]
FM-Trac & \cite{lejeune2020softwarex}  & Python & 3D & Rotation-invariant topology-based matching \\[.75cm]
Part2Track & \cite{janke_part2track_2020} & MATLAB & 2D & Nearest neighbor searching or histogram matching \\[.75cm]
KNOT & \cite{zepeda_o_untying_2021} & Python & 2D \& 3D & Adaptive analysis on the single frame displacements produced from point clouds \\[.75cm]
\textit{\textbf{SerialTrack (this work)}} & \cite{link_SerialTrack}  &  MATLAB & 2D \& 3D & Scale and rotation-invariant topology-based matching and augmented Lagrangian global kinematic compatibility constraint \\ \bottomrule 
\end{tabular}}
\end{table}

\section{Software description}
\label{sec:software_des}
\subsection{Software architecture}
\label{}
\begin{figure}
\centering
\includegraphics[width = 1 \textwidth]{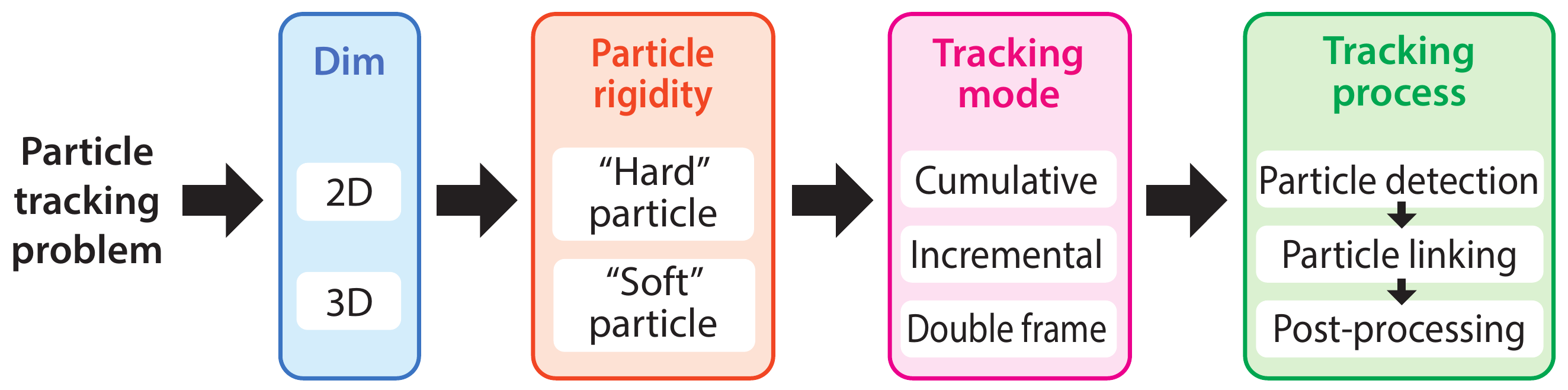}
\caption{Workflow of SerialTrack. The particle tracking setup is most broadly categorized by problem dimensionality as 2D (pixel image) or 3D (volumetric voxel image). Hard vs. soft particles further specify where particle shape warping is included during the solution. The tracking mode defines the scheme for selecting images from the experimental image sequence. Finally, the particles detection, linking, and post-processing strategies are specified given the experimental configuration and desired output data.}
\label{fig:summary}
\end{figure}

The basic workflow of the SerialTrack implementation is summarized in Fig.\,\ref{fig:summary}. 
SerialTrack requires the users to provide their captured 2D and 3D (volumetric) image pairs or image sequence. In solid mechanics and material sciences, 3D image volumes can be scanned in multiple layers using confocal cameras, X-ray CT, MRI, or other imaging modalities. In fluid mechanics, these 3D volumes are typically reconstructed from several camera POVs using multi-view stereoscopy with one or more sensors, see for example \cite{kim_three-dimensional_2016,ardekani_three-dimensional_2013}. 
The code package includes both 2D (SerialTrack\_2) and 3D (SerialTrack\_3) particle tracking problems and three executing modes: (i)\,\textit{incremental}, (ii)\,\textit{cumulative}, and (iii)\,\textit{double frame}. In incremental mode, two consecutive frames are compared to infer incremental motion between frames, while in cumulative mode, later frames are compared with the first, undeformed reference frame to reconstruct their total motions. In 2D cases, we also include the ``double frame'' mode where two frames are taken under every single exposure with a temporal delay and each odd number frame is compared to its subsequent even number frame. One advanced feature of the 2D cumulative mode is that the effect of particle shape distortion can be considered. We consider particles to be ``soft'' if particle shape change coincides with local deformation gradients, and consider particles ``hard'' if the shape is effectively rigid and invariant throughout the experiment.

During the tracking process, we detect each individual particle in both reference and deformed images and then link them to obtain the full-field motions throughout an image sequence. We will discuss these functionalities in Sections\,\ref{sec:detect}-\ref{sec:post}.

\subsection{Software functionalities: particle detection}
\label{sec:detect}
In the SerialTrack implementation, we leverage the state-of-the-art particle detection methods where the minimum size of particles that can be effectively detected are usually 3 pixels by 3 pixels (see more details in the description of each detection method below and references therein). 
In the first of two approaches to detect individual single particles, we employ the same method used in Patel \textit{et al}. \cite{patel2018rapid} where images are thresholded based on a user-specified cutoff to segment particles from background as binary  connected components, and then rapidly localize particle centroids with sub-pixel accuracy using the radial symmetry method \cite{parthasarathy_rapid_2012,liu_fast_2013}.
In the second method, particles are detected by a Laplacian of Gaussian image filtering technique, followed by a Gaussian interpolation of the particle peak intensities \cite{heyman_tractrac_2019,janke_part2track_2020}. Numerous techniques exist for particle segmentation and centroid localization -- while these two perform well in our test cases, for different imaging modalities or particle types other methods may be more appropriate. Good particle detection, segmentation, and localization are critical for accurate tracking. In this regard, the code is highly extensible, and other algorithms can easily be added since the core algorithm simply expects a list of centroid coordinates for each image as input. 

\subsection{Software functionalities: particle linking}
\label{sec:link}
We describe our particle linking process in this section. We summarize the code outline in Algorithm\,\ref{alg:serial_mpt_hardpar}. A mathematical formulation of the particle tracking problem is summarized in \ref{app:prob_form}, where a regularizer is added to the particle matching optimization functional (\ref{eq:PT_cost_func_rg}) to enforce the uniqueness of the solution. In our implementation, this optimization problem is solved by the alternating direction method of multipliers (ADMM) scheme summarized in \ref{app:admm}. To solve the ADMM local step of Eq\,(\ref{eq:al_subpb1}), particles are linked between frames using a new particle descriptor (i.e., feature vector), which is defined based on the topological arrangement of randomly located neighbors in a framework similar to those described in Patel \textit{et al}. \cite{patel2018rapid}, Janke \textit{et al}. \cite{janke_part2track_2020}, and Lejeune \textit{et al}. \cite{lejeune2020softwarex}. Here we improve the previous topology-based relative neighbor feature to be 2D/3D scale- and rotation-invariant, as shown in Section\,\ref{sec:local_linking}. 

During the iterative matching process, a universal outlier removal scheme \cite{ewers_single-particle_2005} is also incorporated to enhance the quality of the reconstructed displacement field. During each ADMM iteration, we also detect and remove ghost particles, i.e., particles that are only detected once in the two compared frames, to improve the robustness of each particle's topology-based feature vector (see Section\,\ref{sec:ghost_rm}). 

To track total, cumulative displacement fields, sometimes called Lagrangian particle tracking, we describe two strategies in Section\,\ref{sec:cum_disp}. 
Finally, we also discuss minimizing errors due to particle shape distortion in Section\,\ref{sec:shape_distortion}, since in some experimental instances particles may deform significantly according to local deformation gradients, and thus shape change must be accounted for in the localization process.

\subsubsection{Topology-based scale and rotation-invariant particle descriptor}
\label{sec:local_linking}
\begin{figure}
\centering
\includegraphics[width = 1 \textwidth]{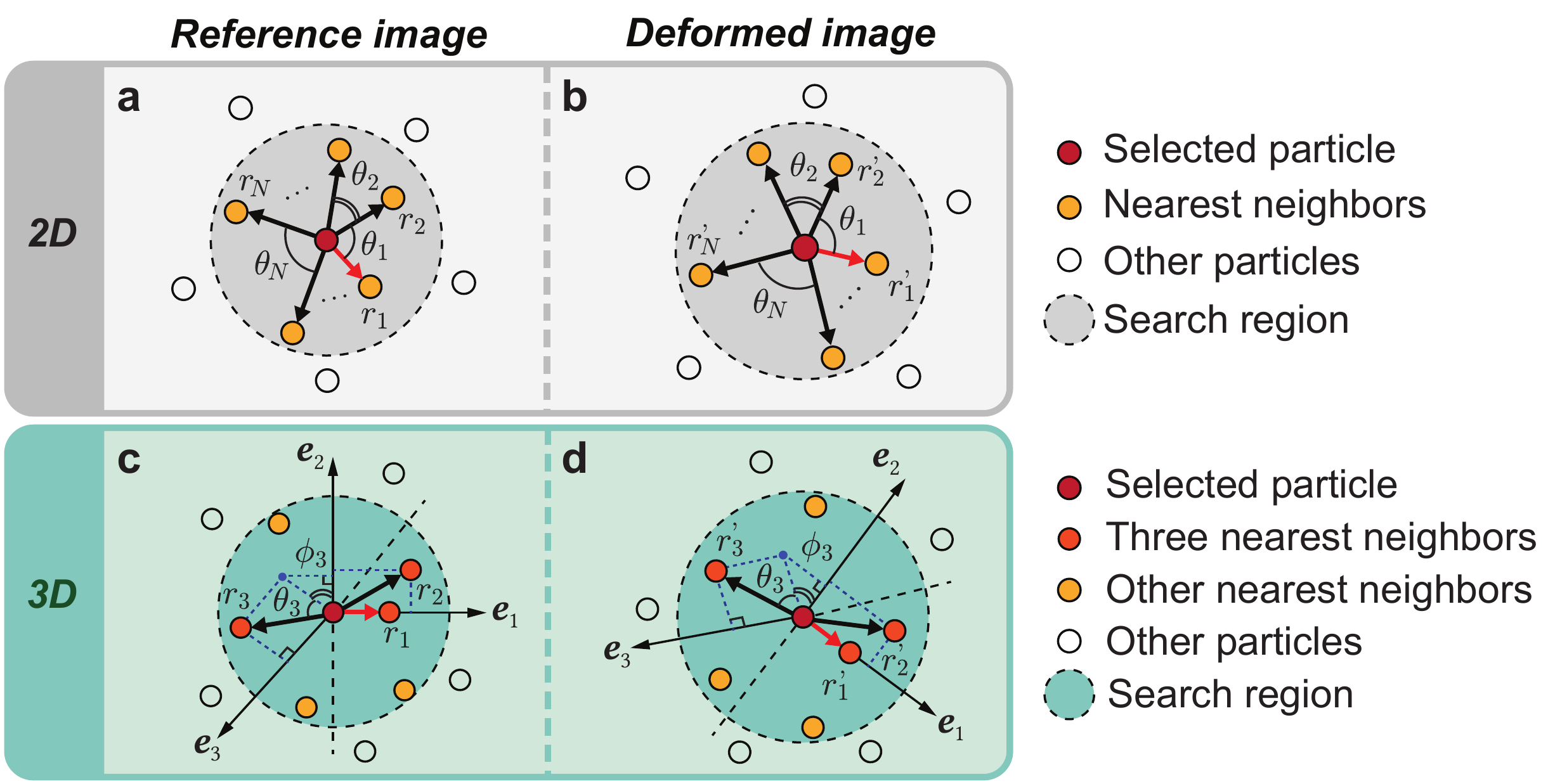}
\caption{A diagram outlining the descriptor generation process. (a) The $k$ radii and angles to nearest neighbor particles within the search distance for each particle (b) The same computation is performed in the deformed image. Simultaneously minimizing Euclidean distance for angular and distance mismatch, we achieve a fast linking that is scale and rotation invariant, and thus robust under most kinematically admissible deformations. (c \& d) The analogous process for a 3D (volumetric) case, where the basis space for the descriptors is computed locally from each particle's three nearest neighbors.}
\label{fig:topology_feature}
\end{figure}

For each individual particle, the relative position between the $k$ nearest neighbor particles and the selected particle is encoded into a complete particle descriptor consisting of two feature vectors. For the 2D case, an angle-based feature is defined as an array of polar angles between each of the $k$ neighboring particles, i.e., $[\theta_1, \theta_2, \cdots, \theta_k]^{T}$,  as shown in Fig.\,\ref{fig:topology_feature}(a-b). An array of interparticle Euclidean distances is also constructed as a distance-based feature, i.e., $[r_1, r_2, \cdots, r_k]^{T}$, where distances are normalized by the first nearest neighbor particle distance. For the 3D case, analogous to the 2D descriptor,  3D radial distances ($r$), polar angles ($\theta$), and azimuthal angles ($\phi$) of the stored $k$ neighboring particles are used to construct particle descriptors for each particle. To establish a coordinate system, $\lbrace \mathbf{e}_1, \mathbf{e}_2, \mathbf{e}_3 \rbrace$, for each particle, we define the first nearest neighbor particle direction as the $\mathbf{e}_1$ axis. The $\mathbf{e}_3$ direction is defined to be perpendicular to the first and second nearest neighbor particles, and must satisfy $\mathbf{e}_3$\,$ \cdot$\,$ \mathbf{r}_3$\,$ >$\,$ 0 $ where $\mathbf{r}_3$ is the third nearest neighbor particle direction. The $\mathbf{e}_2$ axis is defined as $\mathbf{e}_3$\,$ \times$\,$ \mathbf{e}_1$ where $"\times"$ is the cross product. As in 2D, the radial distance ($r$) feature is the Euclidean interparticle distance normalized by the first nearest neighbor particle distance.
The design of these descriptors is advantageous since they fully encode the relative spatial positions of neighboring particles. They can also be cheaper to compute compared to normalized correlation-based tracking algorithms, with a possible the computational cost reduction on the order of (\# of image pixels)\,/\,(\# of particles).  The constructed particle descriptors are scale and rotation invariant, which allows for large deformations and rotations while retaining similarity between descriptors during tracking. 

To establish particle matches between frames, the topology-based descriptor for each identified particle is computed that independently minimizes the Euclidean distances (summation of squared differences) between the distance-based feature and angle-based features, respectively. We consider two particles to be \textit{matched}, i.e., the same physical particle in both frames, if they attain the minimum radial and angular descriptors simultaneously. During ADMM iterations, we apply a particle count scaling strategy, such that the number of nearest neighboring particles for each local matching step, $k$, is exponentially decreased from a user-selected starting value (typically 10's of particles) to 1. When $k$ equals 1, our method is identical to the nearest neighbor search \cite{janke_part2track_2020}. This scaling strategy helps to address variable particle density -- we have found that features relying on many local particles (high $k$) perform well for densely seeded regions, whereas features with lower $k$ are more performant for sparsely seeded regions. In both cases, a maximum search radius for particles to include can be specified to reduce computational cost while building features. 

\begin{algorithm}[!t]
\small
\SetAlgoLined
\KwIn{Two images $f_n$ and $f_{n+t}$}
\textbf{Step 1: }
Detect particles in images $f_n$ and $f_{n+t}$\;
\textbf{Step 2: }
(Optional) Warp detected particle coordinates in $f_{n+t}$ with a predictor for displacement $\hat{\mathbf{u}}$\;
\textbf{Step 3: }
Set dual variable $\boldsymbol{\theta}$ to be zero. Set $IterNum=1$, and  $M=0$\;
\While{ $ \big| \mathbf{\hat{u}}^{k+1}-\mathbf{\hat{u}}^{k} \big| > \varepsilon  $ \textbf{and} $IterNum<IterMax$ \textbf{and} $M < 5$ }{
\%\%\%\%\% {\textbf{Subproblem 1}: fix $\hat{\mathbf{u}}^{k}$ and solve  $\mathbf{u}^{k+1}$ locally} \%\%\%\%\% \;
\textbf{Step 4: } Build a topology-based, scale and rotation invariant descriptor for each particle\;
\textbf{Step 5: } Match particle features and calculate $MatchRatio$\;
\%\%\%\%\% {\textbf{Subproblem 2}: fix ${\mathbf{u}}^{k+1}$ and solve $\hat{\mathbf{u}}^{k+1}$ globally} \%\%\%\%\% \;
\textbf{Step 6: } Solve Eq\,(\ref{eq:al_subpb2})\; 
\textbf{Step 7: } (Optional) Remove ghost particles using Eq\,(\ref{eq:rm_ghost_particles})\;
\textbf{Step 8: } Update dual variable: $\boldsymbol{\theta}^{k+1}$ $\leftarrow$ $\boldsymbol{\theta}^{k}+\hat{\mathbf{u}}^{k+1}-\mathbf{u}^{k+1}$\;
\If{MatchRatio $==$ 1}{$M \leftarrow M+1 $ ($M$: count of ``\textit{MatchRatio} $==$ 1") }} 
\KwOut{Deformation displacement fields from image $f_n$ to $f_{n+t}$: locally solved displacement $\mathbf{u}$ and final global, kinematically compatible   $\hat{\mathbf{u}}$ }  
\caption{Outline of the ``hard'' particle tracking procedure \label{alg:serial_mpt_hardpar}}
\end{algorithm}

\subsubsection{Removing ghost particles} 
\label{sec:ghost_rm}
There may exist particles that are only detected once in the two frames, which are termed ghost particles \cite{schanz_shake--box_2016}. These may occur when part of the sample moves out of the field of view, or due to experimental noise or occlusion. The existence of ghost particles has a two-fold effect on the local topology-based matching: the descriptor built in one frame for a ghost particle will not have a corresponding correct match, and feature descriptors of other neighboring particles omit missing particles. Both effects have a deleterious effect on the accuracy of particle linking and tracking. Therefore after solving each ADMM iteration global step of Eq\,(\ref{eq:al_subpb2}), we attempt to detect and remove ghost particles and re-collect the centroid locations of correctly detected particles in both frames using the following criteria:
\begin{equation}
    \begin{aligned}
        \mathbb{P}_{n} & \leftarrow \bigcup_{P_{n}}\left\lbrace \Big( \min_{P_{n+t} \in \mathbb{P}_{n+t}} |P_{n} - P_{n+t}(\hat{\mathbf{u}})| \Big) < \varepsilon_d \right\rbrace,  \\
        \mathbb{P}_{n+t} & \leftarrow \bigcup_{P_{n+t}}\left\lbrace \Big( \min_{P_{n} \in \mathbb{P}_{n}} |P_{n} - P_{n+t}(\hat{\mathbf{u}})| \Big) < \varepsilon_d \right\rbrace,
    \end{aligned} \label{eq:rm_ghost_particles}
\end{equation}
where $\hat{\mathbf{u}}$ is the current, solved displacement field after the ADMM global step; $\varepsilon_d$ is a user specified critical distance for a detected particle in one frame (i.e., frame $n$) to be diagnosed as a ``ghost particle'' if there does not exist any particle in the other comparing frame (frame $n+t$) within $\varepsilon_d$. All the detected ghost particles will be removed and remaining particles will form the updated particle collections $\mathbb{P}_{n}$ and $\mathbb{P}_{n+t}$ for frame $n$ and frame $n+t$, respectively.

\subsubsection{Tracking cumulative displacement}
\label{sec:cum_disp}
Two strategies are provided by the SerialTrack to reconstruct total, cumulative displacements at each step in an image sequence - namely both cumulative and incremental modes of tracking.  
In cumulative mode, subsequent frames are always compared to the fixed, undeformed reference frame, where total cumulative displacement fields can be obtained directly. For large deformations, we employ the tracked results from the previous frames and leverage a machine learning method to estimate a  displacement predictor  to further improve the tracking accuracy of subsequent frames \cite{yang2021staq}.  In incremental mode, each frame is compared to its next frame. The tracked incremental displacement trajectory segments from each image pair in the sequence can be merged to compute the final cumulative displacements at each time step (see Section\,\ref{sec:post} for more details).

\subsubsection{Effect of particle shape distortion}
\label{sec:shape_distortion}
We optionally consider the effect of particle shape distortion, e.g., for painted circular speckle dots on a 2D sample surface that deform into ellipses during a uniaxial compression/tension test,  which can degrade the particle detection and decrease the tracking accuracy. To distinguish from Algorithm\,\ref{alg:serial_mpt_hardpar} where particles are called ``hard'' and whose shapes are assumed to be rigid, we call these particles ``soft'' and assume their shape distortion coincides with their local, underlying deformation gradients. A modified algorithm to better track these distorted particles is summarized in Algorithm\,\ref{alg:serial_mpt_softpar}. Different from ``hard'' particle tracking, in the ``soft'' particle tracking algorithm, particle centroid locations need to be re-detected using warped images during the ADMM iterations (see Step 4 in Algorithm\,\ref{alg:serial_mpt_softpar}).  

\begin{algorithm}[!t]
\small
\SetAlgoLined
\KwIn{Two consecutive images $f_n$ and $f_{n+t}$}
\textbf{Step 1: }
Detect particles in reference image $f_{n}$\;
\textbf{Step 2: }
(Optional) Predict a displacement predictor\;
\textbf{Step 3: }
Set dual variable $\boldsymbol{\theta}$ to be zero. Set $IterNum=1$, and $M=0$\;
\While{ $ \big| \mathbf{\hat{u}}^{k+1}-\mathbf{\hat{u}}^{k} \big| > \varepsilon  $ \textbf{and} $IterNum<IterMax$ \textbf{and} $M < 5$ }{
\textbf{Step 4: }
Warp image $f_{n+t}$ with current displacement $\hat{\mathbf{u}}^{k}$  and detect particles in images $f_{n+t}$\;
\%\%\%\%\% {\textbf{Subproblem 1}: fix $\hat{\mathbf{u}}^{k}$ and solve  $\mathbf{u}^{k+1}$ locally} \%\%\%\%\% \;
\textbf{Step 5: } Build a topology-based, scale and rotation invariant feature for each particle\;
\textbf{Step 6: }  Match particle features and calculate $MatchRatio$\;
\%\%\%\%\% {\textbf{Subproblem 2}: fix ${\mathbf{u}}^{k+1}$ and solve $\hat{\mathbf{u}}^{k+1}$ globally} \%\%\%\%\% \;
\textbf{Step 7: } Solve Eq.\,(\ref{eq:al_subpb2}) for $\mathbf{\hat{u}}^{k+1}$\; 
\textbf{Step 8: } (Optional) Remove ghost particles using Eq.\,(\ref{eq:rm_ghost_particles})\;
\textbf{Step 9: } Update dual variable: $\boldsymbol{\theta}^{k+1}$ $\leftarrow$ $\boldsymbol{\theta}^{k}+\hat{\mathbf{u}}^{k+1}-\mathbf{u}^{k+1}$\;
\If{MatchRatio $==$ 1}{$M \leftarrow M+1 $ ($M$: count of ``\textit{MatchRatio} $==$ 1")  } }
\KwOut{Deformation displacement fields from image $f_n$ to $f_{n+t}$: locally solved displacement $\mathbf{u}$ and final global, kinematically compatible   $\hat{\mathbf{u}}$ }  
\caption{Outline of the ``soft'' particle tracking procedure \label{alg:serial_mpt_softpar}}
\end{algorithm}

\subsection{Software Functionalities: Post-processing}
\label{sec:post}
After all particles have been tracked, we provide post-processing functions to interpolate tracked displacement fields onto a regular gridded mesh to calculate deformation gradients and strain fields \cite{hazlett2020epifluorescence}. In the incremental tracking mode, the direct, tracked displacement field for each frame is in its current deformed configuration, or in an Eulerian coordinate frame \cite{landauer_qDIC_2018}. Alternatively, we also include post-processing to determine individual particle trajectories throughout the image sequence in a Lagrangian coordinate frame. Particles are linked and all the displacement trajectories are merged  \cite{janke_part2track_2020,novara_particle-tracking_2013} to obtain total particle displacements at each frame for each identified particle. To further improve the cumulative tracking ratios, all the tracked incremental displacement trajectory segments are extrapolated both \textit{before} the segment starting time point  and \textit{after} the segment ending time point. Then we find and join trajectory segments at corresponding time points that are likely from the same particles, which can be merged together. We perform this ``extrapolation and finding'' scheme three to five times or until we do not find un-merged trajectory segments.

%


\section{Illustrative examples}
\label{sec:demo}


\begin{table}[t!] 
\centering
\caption{SerialTrack code parameters for demonstrated illustrative examples for synthetic (``syn'') and experimental ``exp'') test cases }
\label{tab:demo_para}
\resizebox{1\textwidth}{!}{
\begin{tabular}{p{3cm}p{.7cm}p{.7cm}p{.7cm}p{1.3cm}p{1.1cm}p{1.5cm}p{1.1cm}p{1.1cm}p{1.1cm}}
\toprule
Illustrative \qquad \qquad examples & Syn or Exp & Fig. & Dim. & Particle rigidity & Track- ing mode & Bead intensity threshold & Bead radius (p\_size) & Max neighbor\,\# & Size of search field  \\ \midrule
Translation & Syn & \ref{fig:results_syn}a & 2D & Hard & inc & 0.5 & 3 & 25 & Inf\\
Rotation & Syn & \ref{fig:results_syn}b  & 2D & Hard & inc & 0.5 & 3 & 25 & Inf\\
Uniaxial stretch & Syn &  \ref{fig:results_syn}c  & 2D & Hard & cum & 0.5 & 3 & 25 & 50\\
Simple shear & Syn & \ref{fig:results_syn}d & 2D & Hard & cum & 0.5 & 3 & 25 & 50\\ 
Translation & Syn &  \ref{fig:results_syn}e  & 3D & Hard & inc & 0.5 & 3 & 25 & Inf\\
Rotation & Syn &  \ref{fig:results_syn}f  & 3D & Hard & inc & 0.5 & 3 & 25 & Inf\\
Uniaxial stretch & Syn &  \ref{fig:results_syn}g  & 3D & Hard & cum & 0.5 & 3 & 25 & 50\\
Simple shear & Syn &  \ref{fig:results_syn}h  & 3D & Hard & cum & 0.5 & 3 & 25 & 50\\
DIC Challenge v2 & Syn & \ref{fig:results_syn}i-j  & 2D & Hard & inc & 0.5 & 3 & 25 & 50 \\
Uniaxial stretch & Syn &  S3c  & 2D & Soft & cum & 0.5 & 3 & 25 & 50\\
Simple shear & Syn &  S3d & 2D & Soft & cum & 0.5 & 3 & 25 & 50\\
Inertial cavitation & Exp &  \ref{fig:results_exp_2d}a  & 2D & Hard & inc & 0.5 & 2 & 10 & 30\\
Pipe flow & Exp &  \ref{fig:results_exp_2d}b  & 2D & Hard & inc & 0.4 & 2 & 25 & 50\\
Foam compression & Exp &  \ref{fig:results_soft_par}  & 2D & Soft & cum & 0.5 & 3 & 25 & 50\\
Hydrogel shear & Exp &  \ref{fig:results_exp_3d}a  & 3D & Hard & inc & 0.1 & 20 & 5 & 700 \\
Gel indentation & Exp &  \ref{fig:results_exp_3d}b  & 3D & Hard & inc & 0.1 & 3 & 25 & 50\\
\bottomrule 
\end{tabular}}
\end{table}

We assess the SerialTrack method with both synthetic and experimental data sets at various particle seeding densities, as shown in Figs.\,\ref{fig:results_syn}-\ref{fig:results_exp_3d}. Although the synthetic image generation model may not capture all noise and error sources present in experimental images, it serves as a verification and validation for the algorithm where direct, quantitative error measurements can be made between tracking results and ground truth data. Several experimental test cases demonstrate the applicability of this technique for permutations of seeding density and dimensionality, and for both fluid and solid materials. All the used code parameters are summarized in Table\,\ref{tab:demo_para}. A permanent copy of the datasets for the examples can be found on the MINDS@UW open access institutional data repository\footnote{\url{https://minds.wisconsin.edu/handle/1793/82901}}. 

\subsection{Synthetic examples}
\label{sec:syn_eg}
\begin{figure}[h!]
\centering
\includegraphics[width = 1 \textwidth]{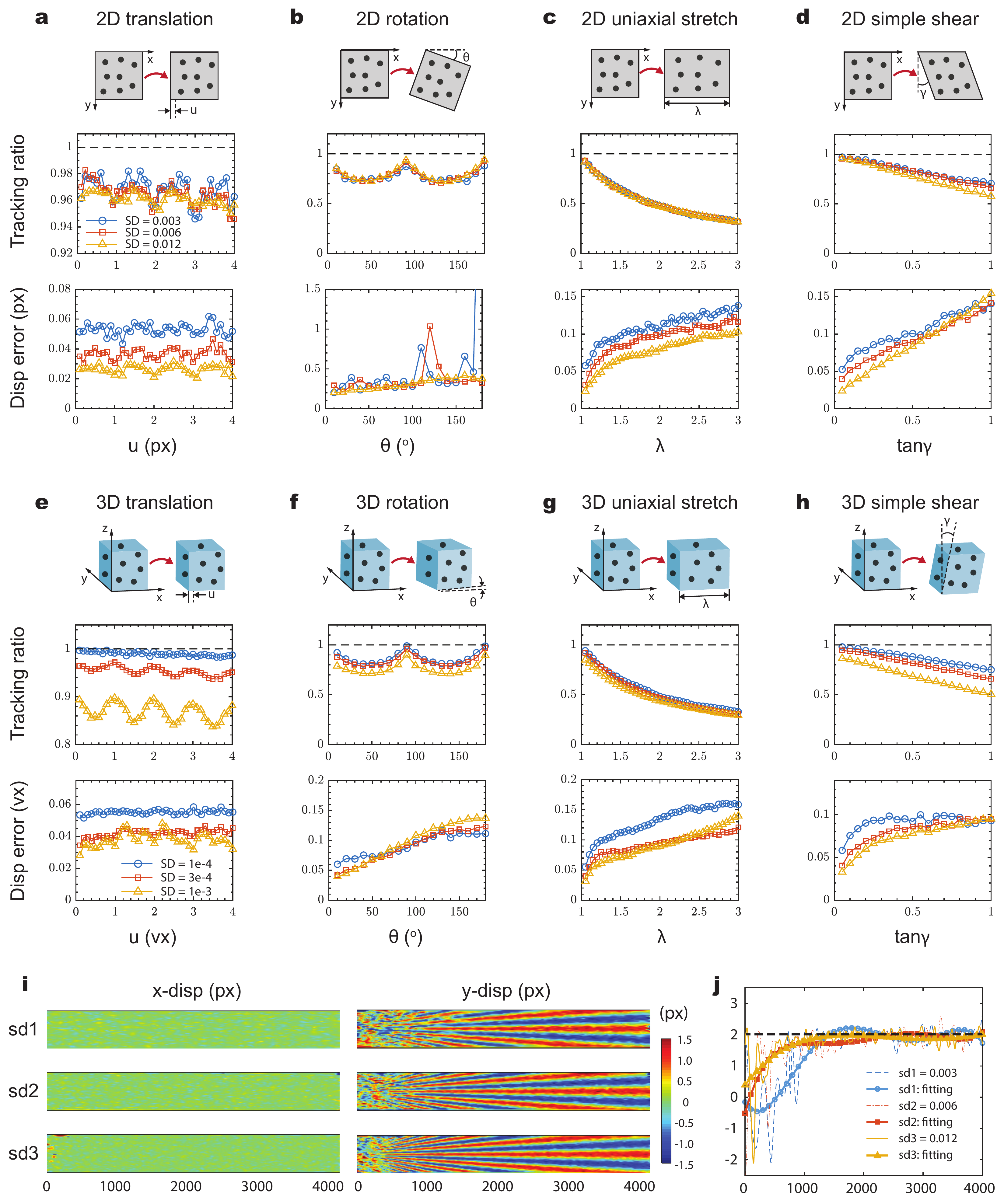}
\caption{The overall ratio of successfully tracked particles to total detected particles as a function of increasing applied motion, i.e., image step, and the displacement RMS error at each step for synthetic 2D (top row) \& 3D (middle row) homogeneous deformations for the proposed SerialTrack method. (SD: seeding density). A synthetic 2D example using a ``star" pattern heterogeneous displacement field is shown at the bottom row.  (a,e)\,Unidirectional translations in the \textit{x}-direction from 0\,pixels (or voxels) to 4\,pixels (or voxels). (b,f)\,Rigid body rotation about the \textit{z}-axis with rotation angles from 0\,$^{\circ}$ to 180\,$^{\circ}$. (c,g)\,Uniaxial stretch in the \textit{x}-direction with stretch ratios from 1 to 3. (d,h)\,Simple \textit{xy}- and \textit{xz}-shear with shear angles from 0\,$^{\circ}$ to 45\,$^{\circ}$. (i)\,Tracked $x$ and $y$ displacement fields in the synthetic ``star" pattern heterogeneous field and their vertical displacements retrieved along the center row are further summarized in (j).}
\label{fig:results_syn}
\end{figure}
For both 2D and 3D synthetic test cases, we generate synthetic images with bead patterns following the steps described in \ref{app:syn}, and their quantitative evaluation results in Fig.\,\ref{fig:results_syn} are shown using tracking ratios and root-mean-squared displacement error. As a baseline, we applied a small first-order motion field in the form of rigid body translation in the \textit{x}-direction from 0 pixels to 4 pixels in 0.1 pixel increments using cumulative tracking. The typical tracking ratios in the 2D and 3D rigid translation are above 95\,\% of particles tracked and 85\,\% of particles tracked, respectively. Displacement root mean square (RMS) errors were computed on the measured centroid locations of each Lagrangian-tracked particle on the order of $10^{-2}$\,pixels. This error level is comparable to the particle localization uncertainty \cite{liu_fast_2013,parthasarathy_rapid_2012}. The deformation gradient tensor, $\mathbf{F}$, of any homogeneous deformation follows the multiplicative decomposition $\mathbf{F} = \mathbf{R} \cdot \mathbf{U} = \mathbf{V} \cdot \mathbf{R}$ where $\mathbf{U}$ and $\mathbf{V}$ are the right and left stretch tensors and $\mathbf{R}$ is a pure rotation of the polar decomposition. Therefore, we include test cases for large, rigid body rotations (rotation angle $\theta$ from 0\,$^{\circ}$ to 180\,$^{\circ}$ in increments of 10\,$^{\circ}$ ), finite uniaxial stretches (stretch ratios $\lambda$ from 1 to 3 in increments of 0.1 and simple shear (shear angle $\gamma$ from 0\,$^{\circ}$ to 45\,$^{\circ}$ in tan($\gamma$) increments of 0.05), as shown in Fig.\,\ref{fig:results_syn}(b-d) for 2D cases and (f-h) for 3D cases. Particle seeding densities (SD) range from 0.003 particles per pixel to 0.012 particles per pixel in 2D synthetic cases, and $10^{-4}$ particles per voxel to $10^{-3}$ particles per voxel in 3D synthetic cases. PTV methods based on an underlying rectilinear grid (typical for image-based measurements) are often challenged by large rotation angles. For large stretches and shears where motions are greater than the inter-particle spacing, local algorithms, such as \textit{k}NN or relaxation methods, non-uniqueness of particle identification leads to poor tracking ratios \cite{patel2018rapid}. In the SerialTrack method, the region-based formulation is designed to minimize these effects and is hybridized with a global optimization that ensures uniqueness and kinematic admissibility of the reconstructed motion field, and thus the tracking ratios remain high and RMS displacement errors are typically below $O$(0.1)\,px.  In all cases, the tracking ratio decrease in part reflects that particles are moving out of the field of view in our referential Eulerian frame. This is most clearly illustrated in the rotation case, where the lowest tracking ratios correspond to approximately 45\,$^{\circ}$ rotation, where the overlap between reference and deformed configuration frame is likewise minimum. For a summary of overall detection ratio and strain RMS error, see SI Fig.\,S1 and Fig.\,S2 for synthetic 2D and 3D results, respectively. Our typical particle tracking applications use rigid particles, however deformable or modulus-matched particles can be important to preserve the verisimilitude of the instrumented test. The method, therefore, also optionally accounts for particle shape distortion. A similar summary of the tracking results for this subclass of reconstructions is given in SI Fig.\,S3. 

Inspired by Reu et al. \cite{dicchallenge2exme}, we designed a synthetic 2D example to test the spatial resolution of the SerialTrack algorithm using a ``star'' pattern displacement field. Both reference and deformed images are $4001$\,pixels $\times 501$\,pixels. The vertical displacement has a varying spatial period, $\lambda$, from 10\,pixels at $x = 1$\,pixels, to 300\,pixels at $x = 4001$\,pixels  according to Eq\,(\ref{eq:dic_challenge_2_period}). The spatial period of the vertical sine wave in the assigned displacement field is proportional to the image position across the width of the image. The magnitude of the periodic vertical displacements is $\pm$\,2 pixels.  The ground truth of horizontal displacement is zero. We test different bead seeding densities (SD) of (0.003, 0.006, and 0.012) beads per pixel.  No additional noise is added during the image generation. 
\begin{equation}
    \lambda = 10 + \frac{300-10}{4001} (x-1) \label{eq:dic_challenge_2_period}
\end{equation}

Both SerialTrack tracked horizontal and vertical displacement fields are summarized in Fig.\,\ref{fig:results_syn}(i).
Vertical displacements retrieved along the center row ($y=251$) are further summarized in Fig.\,\ref{fig:results_syn}(j). The ground truth of the vertical displacement at $y=251$\,pixels is 2\,pixels. First, we find that SerialTrack can resolve heterogeneous deformations well. Particularly, using dense particles where SD$>$0.006 particles per pixel can accurately recover highly oscillating deformation fields when $x>500$\,pixels, where the wave length is greater than 46\,pixels. We also find that the tracked vertical displacement for highly oscillating deformation on the left side is underestimated and noisy. This is not surprising, because with a topology-based particle tracking approach it is challenging to resolve heterogeneous deformation whose characteristic wavelength ($\lambda$) is smaller than half of the averaged nearest-neighbor-particle distance $\sim O(\text{SD}^{-1/2})$. Compared with other subset-based correlation methods \cite{dicchallenge2exme}, the SerialTrack method is not only able to solve dense particles but also sparse particles. Additionally, it can be computationally cheaper, with a potential computational cost reduction on the order of (\# of image pixels)\,/\,(\# of particles).

\subsection{Experimental examples}
With the synthetic deformation cases showing strong performance across a range of deformation modes and particle densities for both 2D and 3D, we move to a variety of experimental test cases. Test cases are conducted in both 2D and 3D, with sparse and dense particles, and are summarized in Figs.\,\ref{fig:results_exp_2d}-\ref{fig:results_exp_3d}. Additionally, a 2D large deformation uniaxial compression experiment is employed to demonstrate the  ``soft particle'' implementation in Section\,\ref{sec:demo_soft} where the ink-printed circular dots have significant shape distortion.

\subsubsection{2D ``hard'' particle examples}
\begin{figure}[h!]
\centering
\includegraphics[width = 0.9 \textwidth]{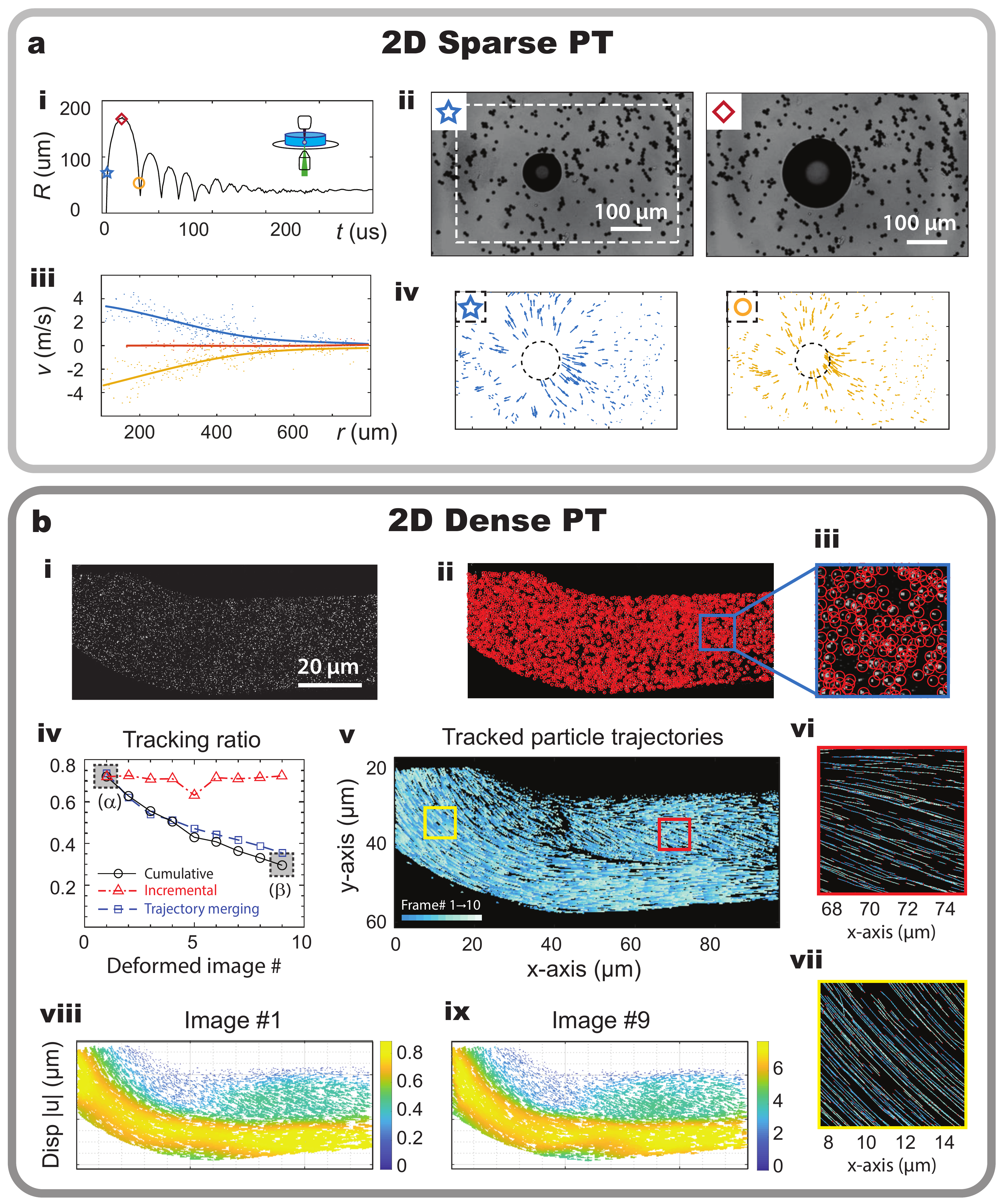}
\caption{ 2D ``hard'' particle tracking experimental examples. (a)\,Tracking laser-induced inertial cavitation in a soft material specimen with sparse particles \cite{McGhee2021-ze}. (a:i)\,Experimentally measured bubble radius vs. time curve. Three specific frames of interest are denoted by a blue star, red diamond, and yellow circle, which corresponds to the expansion, maximum bubble radius, and near the first collapse, respectively. (a:ii)\,Raw high-speed camera frames at the blue star and red diamond time points. (a:iii)\,Reconstructed radial velocity  vs. radial distance curves at the three marked time points in (a:i). (a:iv)\,Reconstructed velocity fields at the blue star and yellow circle time points. (b)\,2D dense particle tracking example of flow through a bent pipe \cite{janke_part2track_2020}. (b:i)\,One typical frame in the time-resolved image sequence. (b:ii-iii)\,Detected single particle centroids using the Laplacian of Gaussian filtering technique are circled in red. (b:iv)\,Particle tracking ratios. (b:v-vii)\,Tracked trajectories. (b:viii-ix)\,Tracked cumulative displacement fields for the first and  ninth frames.   }
\label{fig:results_exp_2d}
\end{figure}

Figure\,\ref{fig:results_exp_2d}a shows an example of 2D sparse particle tracking in a laser-induced cavitation event in a soft material specimen. Following McGhee et al. \cite{McGhee2021-ze},  15\,$\upmu$m polystyrene particles are embedded into a single, flat plane within the bulk of a gelatin hydrogel. Then the sample is placed into the optical path of a laser-based cavitation system \cite{estrada2018jmps,yang2020eml}. A 6\,ns laser pulse is pathed through the backport of an inverted microscope and focused via a 20$\times$ magnification, 0.5 numerical aperture (NA) (i.e., 20$\times$/0.5NA) imaging objective onto the imaging plane to induce a cavitation bubble on the same plane as the embedded particles.
Figure\,\ref{fig:results_exp_2d}a(i) shows the resulting bubble radius vs. time curve with call-outs for specific frames of interest denoted by a blue star, red diamond, and yellow circle which corresponds to the expansion, maximum bubble radius, and near the first collapse, respectively. Two images of a typical cavitation event are shown in Fig.\,\ref{fig:results_exp_2d}a(ii). By tracking motions of the embedded 2D sparse particles in this image sequence, we can reconstruct the evolution of the resulting time-resolved velocity fields. For example,  the radial velocity vs. radial distance  curves are computed and plotted in  Fig.\,\ref{fig:results_exp_2d}a(iii). Velocity fields at the marked expansion (blue star) and collapse (yellow circle) time points in Fig.\,\ref{fig:results_exp_2d}a(i) are summarized in Fig.\,\ref{fig:results_exp_2d}a(iv).

As an example of a 2D dense particle tracking, we examine a case where the data originates from high-speed PIV measurements of flow in a tube \cite{janke_part2track_2020}, as shown in Fig.\,\ref{fig:results_exp_2d}b(i). We use the same Laplacian of Gaussian image filtering technique as described  \cite{janke_part2track_2020} to detect single particles (see Fig.\,\ref{fig:results_exp_2d}b:ii-iii).  We tested both the incremental and cumulative modes (see Fig.\,\ref{fig:results_exp_2d}(b:v-vii)). In the cumulative mode, we directly track the total, cumulative displacement of each individual particle. In the incremental mode, the cumulative displacements are computed by merging trajectory segments (refer to Sec. \ref{sec:post}). The final cumulative tracking ratio is given in each case in Fig.\,\ref{fig:results_exp_2d}(b:iv). The reconstructed cumulative displacement fields for the first and ninth frames are visualized in Fig.\,\ref{fig:results_exp_2d}(b:viii-ix), where the cumulative displacement in the ninth frame is large but still well tracked by the SerialTrack implementation.

\subsubsection{2D ``soft'' particle example} \label{sec:demo_soft}
\begin{figure}[h!]
\centering
\includegraphics[width = 1 \textwidth]{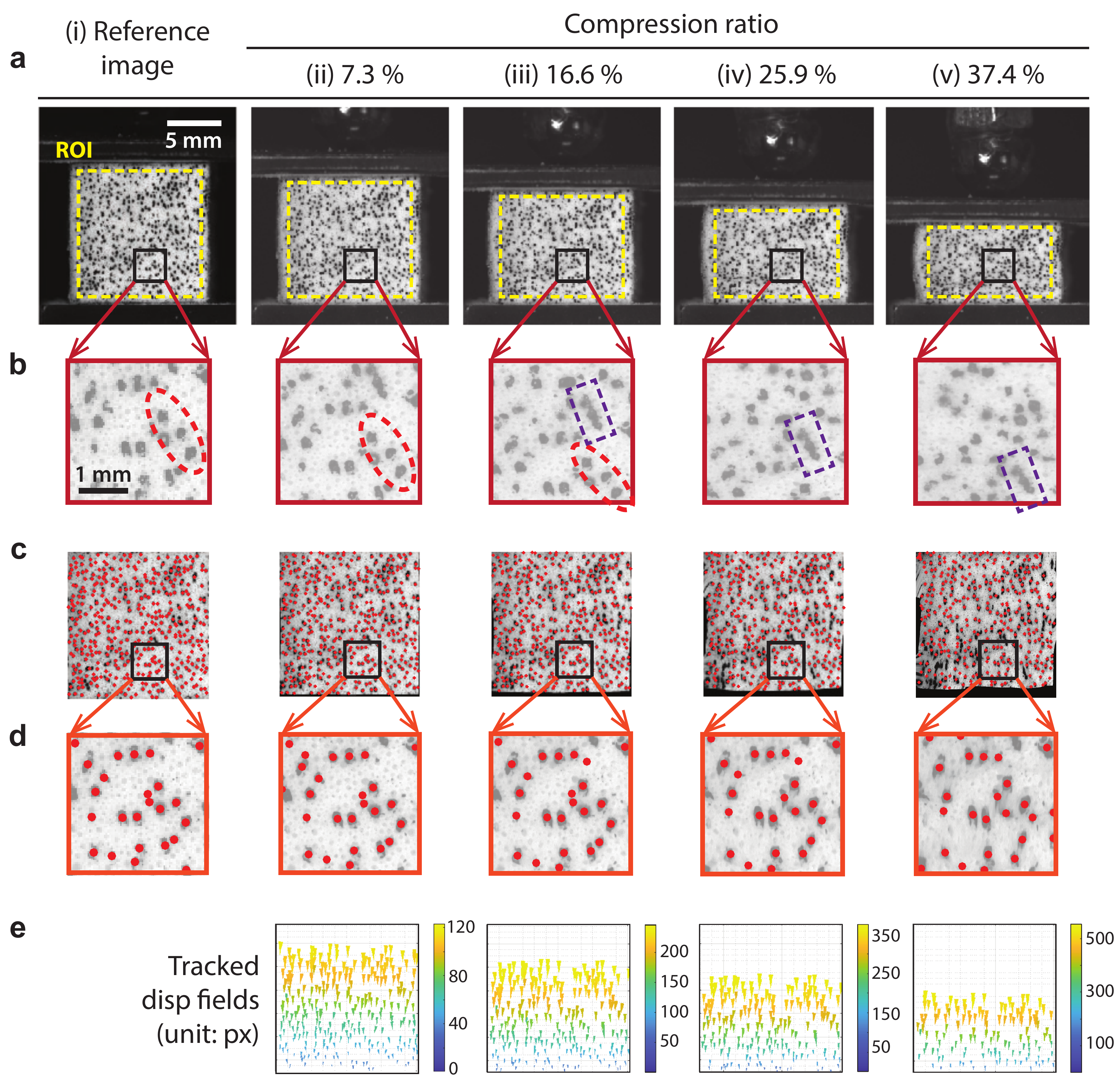}
\caption{Demonstration of the ``soft'' particle tracking (Algorithm\,\ref{alg:serial_mpt_softpar}) via a large deformation uniaxial compression experiment. (a)\,Experimental reference and selected deformed images where regions of interest (ROIs) are marked using dashed line rectangular boxes. Insets are magnified and shown in (b) where three ellipses and three rectangles are marked to highlight the same locations on the surface of the specimen that underwent large deformations, respectively. During the ``soft'' particle tracking process, deformed images are iteratively warped and particles are detected as shown in (c-d). Final tracked cumulative displacement field cone plots are summarized in (e). }
\label{fig:results_soft_par}
\end{figure}

Here we demonstrate the SerialTrack code on a large deformation uniaxial compression experiment on an open-cell polyurethane foam sample with a nominal density of 240\,kg/m$^3$. The dimensions of the foam specimen were approximately 12.7\,mm $\times$ 12.7\,mm $\times$ 12.7\,mm. The experimental setup and other experimental details can be found in Yang \textit{et al}. \cite{yang2021smart}. The reference and deformed images at compression ratios of 7.3\,\%, 16.6\,\%, 25.9\,\%, and 37.4\,\% are shown in Fig.\,\ref{fig:results_soft_par}a, with magnified insets shown in  Fig.\,\ref{fig:results_soft_par}b. Three dashed-line ellipses and three rectangles are marked to highlight the same locations on the front surface of the testing specimen that underwent large deformations.

As described in Algorithm\,\ref{alg:serial_mpt_softpar}, deformed images are iteratively warped and single particles are detected during each ADMM iteration. Here we present the final warped images in Fig.\,\ref{fig:results_soft_par}c. The corresponding detected particle centroids are marked with red dots and shown under magnification in Fig.\,\ref{fig:results_soft_par}d. The final tracked cumulative displacements are visualized in cone plots and summarized in Fig.\,\ref{fig:results_soft_par}e. 
 
\subsubsection{3D examples}
\begin{figure}[h!]
\centering
\includegraphics[width = 0.7 \textwidth]{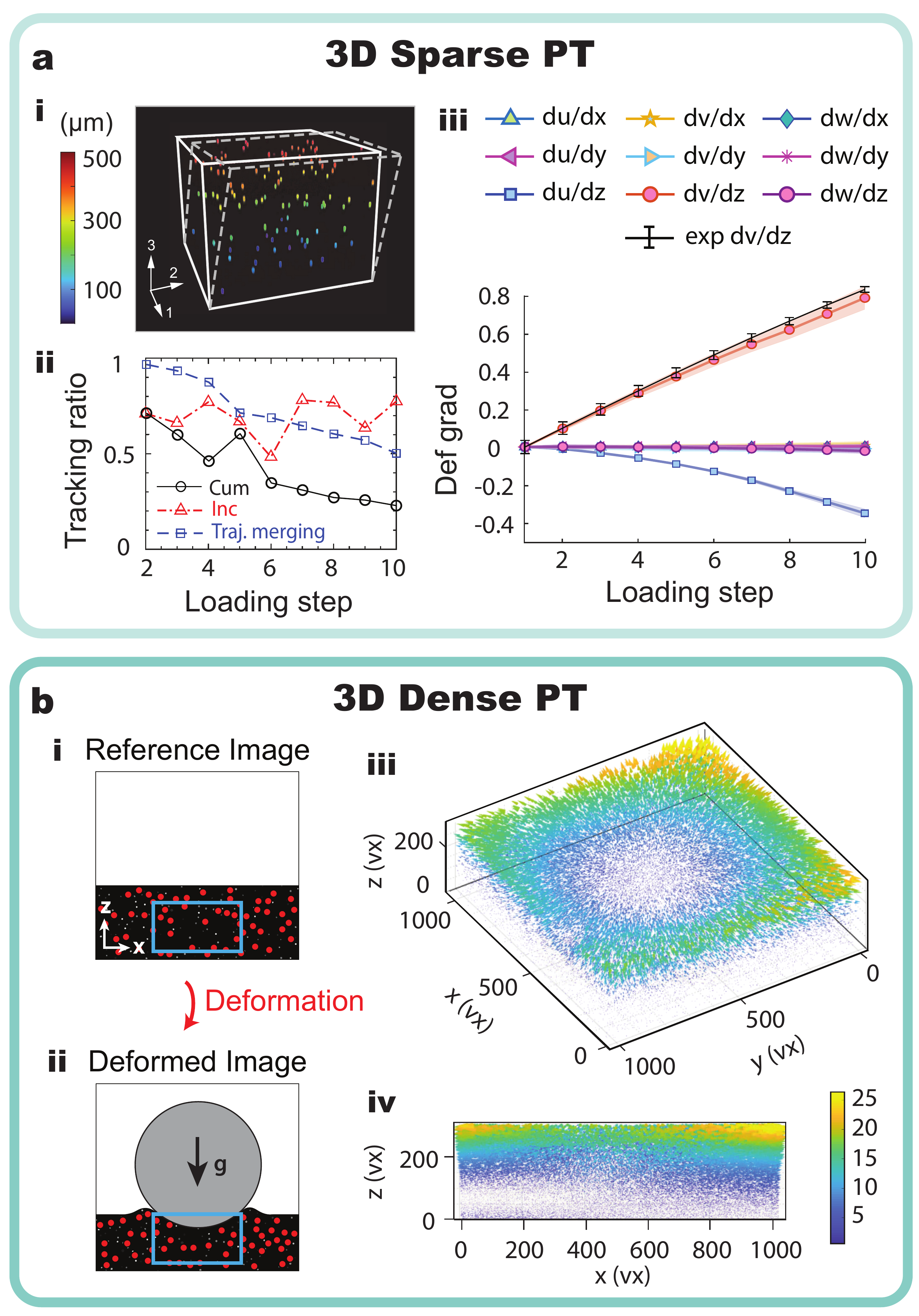}
\caption{Tracking 3D deformations. (a)\,A 3D shear example (du/dz\,$\neq$\,0, dv/dz\,$\neq$\,0) with sparse particles. (a:i)\,The shear deformation was imposed quasi-statically on the hydrogel sample. All the detected particles are colored depending on the z-coordinate of their centroids. (a:ii)\,Tracking ratios. (a:iii)\,Reconstructed deformation gradients, which agree well with the nominal crosshead motion (coded as ``exp''). (b)\,A 3D spherical indentation example where particles are densely seeded. (b:i-ii)\,Sketch of the reference and deformed configurations of the spherical indentation experiment (not to scale). (b:iii-iv)\,3D cone plot and the xz-plane projection of the tracked 3D deformation.  }
\label{fig:results_exp_3d}
\end{figure}

As a sparse 3D tracking example case, inspired by biological applications such as traumatic brain injury \cite{summey_thesis_2020,bar-kochba_strain_2016}, we seeded 5\,$\mu$m fluorescent microparticles at a 1.5\,\% vol/vol fraction in a soft polyacrylamide hydrogel and deformed the hydrogel in a simple-shear-like mode on a confocal laser point scanning microscope using a 20$\times$/0.5NA (approximately 1\,$\upmu$m voxel size) imaging objective. The shear deformation was imposed quasi-statically (1\,minute per step) in 10 steps in nominally 4\,\% engineering shear strain increments from 0\,\% strain to 40\,\% strain, and a total of $O$(100) particles were tracked as shown in Fig.\,\ref{fig:results_exp_3d}a(i) where the artificial color of the particles depends on the z-coordinate.  Results, including the particle tracking ratios, nominal crosshead, and reconstructed deformation gradient tensor components, are shown in Fig.\,\ref{fig:results_exp_3d}a(ii,iii). In addition, note that the final cumulative tracking ratio obtained by merging incrementally tracked trajectory segments is higher than the direct cumulative mode (see Fig.\,\ref{fig:results_exp_3d}a(ii)).

We also test our SerialTrack method for tracking densely seeded, three-dimensional particles. In this experiment, a 1 mm diameter stainless steel sphere with a density of 7750\,kg/m$^3$ was placed onto the surface of a submerged soft polyacrylamide (PA) hydrogel to perform the spherical indentation under the force of gravity ($g$), as shown in Fig.\,\ref{fig:results_exp_3d}b(i-ii). 3D volumetric image stacks (image size: 1024\,voxels $\times$ 1024\,voxels $\times$ 445\,voxels) containing fluorescent beads were scanned before and after the indentation deformation near the hydrogel surface using multiphoton microscopy and a 25$\times$/1.15NA water immersion objective \cite{yang2020augmented}. All other experimental parameters can be found in \cite{yang2020augmented}. The three-dimensional cone plot and the xz-plane projection of the tracked 3D deformation are shown in Fig.\,\ref{fig:results_exp_3d}b(iii) and (iv), respectively.

\section{Impact and conclusions}
\label{sec:impact}
%
%
%
%
%

Particle tracking, often called single particle tracking (SPT) and particle tracking velocimetry (PTV), is of key importance for full-field quantitative analysis of dynamic processes, typically from time-lapse image data. It operates by detecting and tracking individual tracer particles or fiducial markers during a time-resolved image sequence. Here we present a new hybrid local-global tracking algorithm that builds an iterative scale and rotation invariant topology-based feature for each particle within a multi-scale tracking process, where the global kinematic compatibility of the final tracked displacement field is optimized. SerialTrack is able to track particles in both 2D and 3D images with both sparse and dense particle seeding densities, and can accurately reconstruct large, finite deformation and velocity fields. We also consider the effect of shape distortion of particles due to their local deformation gradients. We used synthetic examples to verify and validate the implementation and provide an estimate of the spatial resolution capabilities. We then demonstrate the performance and post-processing routines on a variety of experimental test cases including 2D and 3D, sparse and dense, and soft and hard particles. The provided open-source code package implements the proposed SerialTrack method and allows users to apply the code directly to their own research.


\section{Conflict of interest}
%

No conflict of interest exists:
We confirm that there are no known conflicts of interest associated with this publication and there has been no significant financial support for this work that could have influenced its outcome.

Certain commercial equipment, software and/or materials are identified in this paper in order to adequately specify the experimental procedure. In no case does such identification imply recommendation or endorsement by the National Institute of Standards and Technology, nor does it imply that the equipment and/or materials used are necessarily the best available for the purpose.

\section*{Acknowledgements}
\label{}
The authors thank the U.S. Office of Naval Research for research support under the ``PANTHER'' program, award number N000142112044 through Dr. Timothy Bentley. This research was performed while A.K.L. held an NRC Research Associateship award with Aaron Forster at the National Institute of Standards and Technology.

\appendix

\section{Particle tracking problem formulation}
\label{app:prob_form}
We assume that each image is in a sequence of images from the $0^{th}$ (reference configuration) to $N^{th}$ (final, fully deformed) configuration, where each image is defined by a grayscale intensity field $f_n(\mathbf{x})$, consisting of multiple superposed intensity sources ($\mathcal{N}$, e.g., from fluorescent particles) in the image domain as:
\begin{equation}
    f_n(\mathbf{x}) = \sum_{\mathbf{P} \in \mathbb{P}_n} \mathcal{N}\left(\mathbf{x}; {A}(\mathbf{P}) \exp{ \left(  -\frac{|\mathbf{x}-\mathbf{P}|^2}{2  {\sigma}(\mathbf{P})^2}  \right)\mathbf{I} }\right), 
\end{equation} 
where $\mathbf{x}$ denotes each image pixel; $\mathbf{P}$ is the coordinate of each individual particle centroid; ${A}$ is the maximum particle intensity with ${\sigma(\mathbf{P})}$ the standard deviation of intensity decay for each particle; $\mathbf{I}$ is an identity matrix. $\mathbb{P}_n$ is the collection of all particles in image $f_n$, $C(n)$ is the total number of particles in image $f_n$, and we define the operator $\mathcal{E}$ as the \textit{particle detection} process that extracts the centroid position ($\mathbf{P}_i$) of each particle such that
\begin{equation}
    \mathbb{P}_n = \left\lbrace \mathbf{P}_0, \mathbf{P}_1, \cdots, \mathbf{P}_{C(n)} \right\rbrace := \mathcal{E}(f_n(\mathbf{x}))
\end{equation}

The subsequent \textit{particle linking} process solves for the unknown displacement field $\mathbf{u}_{i}$ from image $n$ in the sequence to a later image $n+t$ in the sequence. For \textit{incremental} mode we set $t \triangleq 1$, and for \textit{cumulative} mode $n \triangleq 0$ and $t \in N$, where $t$ is in general a positive integer number of image frames. The linking process solution minimizes a cost function, for example, in a simple case the sum of squared differences,
\begin{equation}
    \min_{\mathbf{u}_{i}} \int_{\mathbf{x}} {\left| \mathcal{E}(f_n(\mathbf{x})) - \mathcal{E}(f_{n+t}(\mathbf{x}-\mathbf{u}_i))  \right|^2 } \ d \mathbf{x} \label{eq:PT_cost_func}
\end{equation}
%
However, the above optimization problem is ill-posed. In general, the cost function is not convex and the solution is not unique, since the matching of particles $\mathbb{P}_n$ to $\mathbb{P}_{n+t}$ has no guarantee of uniqueness and in practices mismatches and non-matching particle links frequently exist. To reduce displacement noise introduced by mis-linked particles, global regularization penalties are further added to the optimization cost function. This optimization problem can be efficiently solved by the {\it alternating direction method of multipliers} (ADMM) after adding a global slack variable, $\mathbf{\hat{u}_i}$. The modified optimization problem is:
\begin{equation}
\begin{split}
 & \min_{\mathbf{u}_i} \int_{\mathbf{x}} {\left|  \mathcal{E}(f_n(\mathbf{x})) -  \mathcal{E}(f_{n+t}(\mathbf{x}-\mathbf{u}_i))  \right|^2 + \frac{\alpha}{2} \left| \nabla \hat{\mathbf{u}}_i \right|_F^2 } \ d \mathbf{x},  \\
 & \text{subject to } \mathbf{u}_i = \hat{\mathbf{u}}_i 
\end{split} \label{eq:PT_cost_func_rg}
\end{equation}
where the coefficient $\alpha$ is a positive weight of the added regularizer and $|\cdot|_F$ is the Frobenius norm for tensors such that $|\mathbf{A}|_F^2:=\sum_{i} \sum_{j} |A_{ij}|^2 $. 
The first term in (\ref{eq:PT_cost_func_rg}) is the displacement from particle matches using the \textit{local} linking algorithm information, which can be solved quickly and in parallel as discussed in Sect\,\ref{sec:local_linking} for our implementation. The second term in Eq\,(\ref{eq:PT_cost_func_rg}) penalizes \textit{global} displacement variance and noise.
Other global regularization schemes can also be applied if there is additional, \textit{a priori} known information about the physics of the problem.
The combined local and global optimization solution is implemented in an iterative fashion, such that local particle matching is informed by the globally refined displacement field to yield a final unique and kinematically admissible displacement field $\mathbf{u}_i$ with local accuracy and resolution from individually tracked centroid locations.

\section{Alternating direction method of multipliers} 
\label{app:admm}
To efficiently solve the optimization problem posed in Eq.\,(\ref{eq:PT_cost_func_rg}), it can be rewritten for a given displacement step $\mathbf{u}$ in ADMM form as \cite{boyd2011distributed}:
\begin{align}
     \mathcal{L}(\mathbf{u},\mathbf{\hat{u}},\boldsymbol{\theta}) = \int_{\mathbf{x}} {\left|  \mathcal{E}(f_n(\mathbf{x})) - \mathcal{E}(f_{n+t}(\mathbf{x}-\mathbf{u}))  \right|^2 + \frac{\alpha}{2} \left| \nabla \hat{\mathbf{u}} \right|^2 }  +  \frac{\mu}{2} \left| \hat{\mathbf{u}} - \mathbf{u} + \boldsymbol{\theta} \right|^2   \ d \mathbf{x} \label{eq:PT_cost_func_al}
\end{align}
where $\mu$ is a positive coefficient of the added augmented Lagrangian penalty and $\bm{\theta}$ is an introduced dual variable.  
During each ADMM iteration step, we first decompose the global minimization problem Eq.\,(\ref{eq:PT_cost_func_al}) into independent, local problems (see Sect\,\ref{sec:local_linking}), then all the local solutions, $\mathbf{u}$, are projected onto a global, kinematically compatible space where the global auxiliary displacement field $\hat{\mathbf{u}}$ is admissible. Mathematically, given the results $\lbrace \mathbf{u}_i^k \rbrace$, $\lbrace \hat{\mathbf{u}}_i^k \rbrace$, $\lbrace \boldsymbol{\theta}_i^k \rbrace$ in the $k$th step, we solve the $(k+1)$th update using the following steps: \\

\noindent $\bullet$ \ \underline{\textbf{Subproblem 1}: Local update.} While holding $\lbrace \hat{\mathbf{u}}_i^k \rbrace$ and $\lbrace \boldsymbol{\theta}_i^k \rbrace$ fixed, we minimize Eq.\,(\ref{eq:PT_cost_func_al}) over $\lbrace \mathbf{u}_i \rbrace$ to obtain $\lbrace \mathbf{u}_i^{k+1} \rbrace$. Since $\lbrace \hat{\mathbf{u}} \rbrace$ is fixed and $\mu$ can be a small value, this problem is broken down into a series of local problems that can be solved independently using local topology-based feature matching to obtain a displacement guess:
\begin{equation}
    \mathbf{u}^{k+1} := \text{arg}\,\min_{\mathbf{u}} \mathcal{L}(\mathbf{u},\mathbf{\hat{u}}^{k}, \boldsymbol{\theta}^{k}) \label{eq:al_subpb1}
\end{equation}

\noindent $\bullet$ \ \underline{\textbf{Subproblem 2}: Global update.} While holding $\lbrace  {\mathbf{u}}_i^k \rbrace$ and $\lbrace \boldsymbol{\theta}_i^k \rbrace$ fixed, we minimize $\mathcal{L}$ over $\lbrace \hat{\mathbf{u}} \rbrace$ such that
\begin{equation}
    \hat{\mathbf{u}}^{k+1} := \text{arg}\,\min_{\hat{\mathbf{u}}} \mathcal{L}(\mathbf{u}^{k+1},\mathbf{\hat{u}}, \boldsymbol{\theta}^{k}) \label{eq:al_subpb2}
\end{equation}
This is a global problem, but is independent of the original image sequence $\mathbf{f}$ since it only relies on the displacements computed from the local particle linking step. Indeed, it leads to a well-posed linear problem: 
\begin{equation}
    \left( - \frac{\alpha}{\mu} \nabla \cdot \nabla + \mathbf{I} \right) \hat{\mathbf{u}} = \mathbf{u}^{k+1} - \boldsymbol{\theta}^{k}
\end{equation}

\noindent $\bullet$ \ \underline{\textbf{Subproblem 3}: Dual variable update.} We finally update the dual variable $\lbrace \boldsymbol{\theta} \rbrace$ as follows:
\begin{equation}
    \boldsymbol{\theta}^{k+1} := \boldsymbol{\theta}^{k} + \hat{\mathbf{u}}^{k+1} - \mathbf{u}^{k+1}.
\end{equation}

In practice, the smoothing parameter $\alpha$ is carefully chosen in the range $\alpha/\mu = O(10^{-3}) \sim O(10^{-1}) $ based on the expected smoothness of the deformation field. The parameter $\alpha/\mu$ can be further tuned using the L-curve method \cite{hansen1993use}. 

\section{Synthetic image generation}
\label{app:syn}

In each reference image, isolated spherical beads are randomly seeded using a 2D or 3D Gaussian intensity profile as an approximation of a random, isotropic image pattern. A typical Gaussian point spread function (PSF) with amplitude $A$ and spread $\sigma$ and located at $\mathbf{x}$ is expressed as
\begin{equation}
    \text{PSF}(\mathbf{x}) = A \text{exp} \left(  - \sum_{i=1}^{d} \frac{{x}_i^2}{2 \sigma^2}  \right) \mathbf{I}
\end{equation}
where $d$ is the image dimensionality and $x_i$ is the $i^{th}$-component of $\mathbf{x}$; $\mathbf{I}$ is the identity tensor. We choose $\sigma = 1$ to approximate a circular/spherical particle in the volume image with a diameter of approximately 5 pixels or voxels. All the beads are sampled randomly with seeding density SD which denotes the number of particles per pixel or voxel. To avoid particles overlapping in the synthetic images, a Poisson disc sampling algorithm is used to seed center-point locations with a minimum separation distance between particles equal to the particle diameter \cite{patel2018rapid}. The particle positions in the deformed images are calculated via the imposed displacement field and grayscale values are interpolated into the image. In addition, 5\,\% white Gaussian noise has been added to the synthetic images to roughly approximate the experimental noise in our images.

\bibliographystyle{elsarticle-num} 
\bibliography{reference}









\end{document}


\begin{frontmatter}




\title{Supplemental Information for \\
``SerialTrack: ScalE and Rotation Invariant Augmented Lagrangian Particle Tracking''}

\author{Jin Yang \fnref{1}}
\author{Yue Yin \fnref{1,2}}
\author{Alexander K. Landauer \fnref{3}}
\author{Selda Buyuktozturk \fnref{1,4}}
\author{Jing Zhang \fnref{1}}
\author{Luke Summey \fnref{1}}
\author{Alexander McGhee \fnref{1}}
\author{Matt K. Fu \fnref{5}}
\author{John O. Dabiri \fnref{5}}
\author{Christian Franck \corref{cor1}\fnref{1}}
\cortext[cor1]{Corresponding author}
\ead{cfranck@wisc.edu}
\address[1]{University of Wisconsin-Madison, Mechanical Engineering, Madison, WI, USA}
\address[2]{Department of Mechanical Engineering, Carnegie Mellon University, Pittsburgh, PA, USA}
\address[3]{National Institute of Standards and Technology, Gaithersburg, MD, USA}
\address[4]{School of Engineering, Brown University, Providence, RI, USA}
\address[5]{Graduate Aerospace Laboratories, California Institute of Technology, Pasadena, CA, USA}

\end{frontmatter}


\section{2D Synthetic uniform deformation: ``hard'' particles}

\begin{figure}[h!]
\centering
\includegraphics[width = 1 \textwidth]{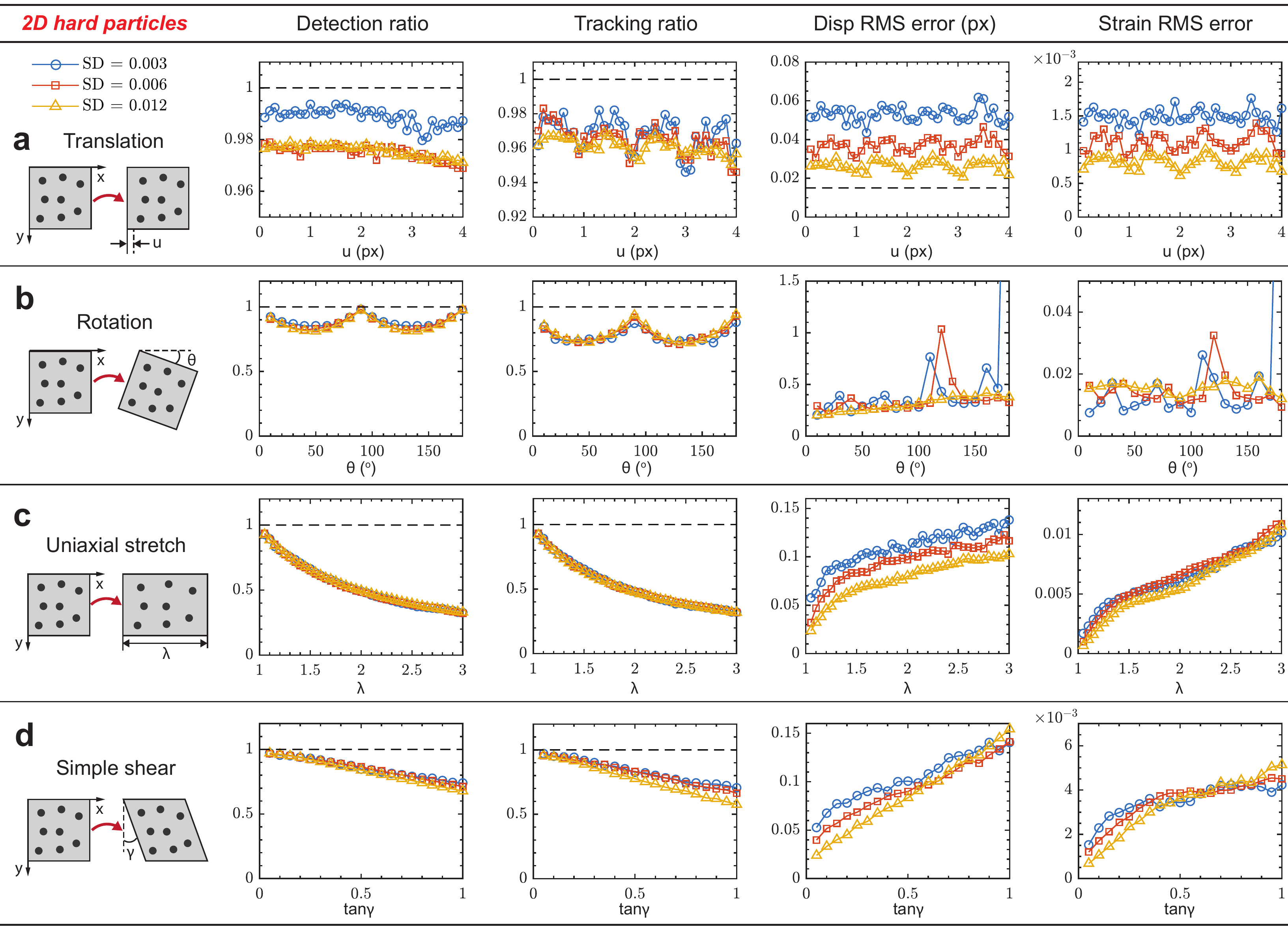}
\caption{The ratio of detected to total ``hard" 2D particles, the overall ratio of successfully tracked particles to total detected particles, and the displacement and strain RMS errors as a function of increasing applied motion, i.e., image step, for synthetic 2D homogeneous deformations for the proposed ALT-SCRIPT method. (a)\,Unidirectional translations in the x-direction from 0 to 4 pixels.
(b)\,Rigid body rotation about the $z$-axis with rotation angles from $0^{\circ}$ to $180^{\circ}$. (c)\,Uniaxial stretch in the $x$-direction with stretch ratios from 1 to 3. (d)\,Simple $xy$-shear with shear angles from $0^{\circ}$ to $45^{\circ}$.}
\label{fig:si_2d_syn}
\end{figure}

\newpage
\section{3D Synthetic uniform deformation: ``hard'' particles}

\begin{figure}[h!]
\centering
\includegraphics[width = 1 \textwidth]{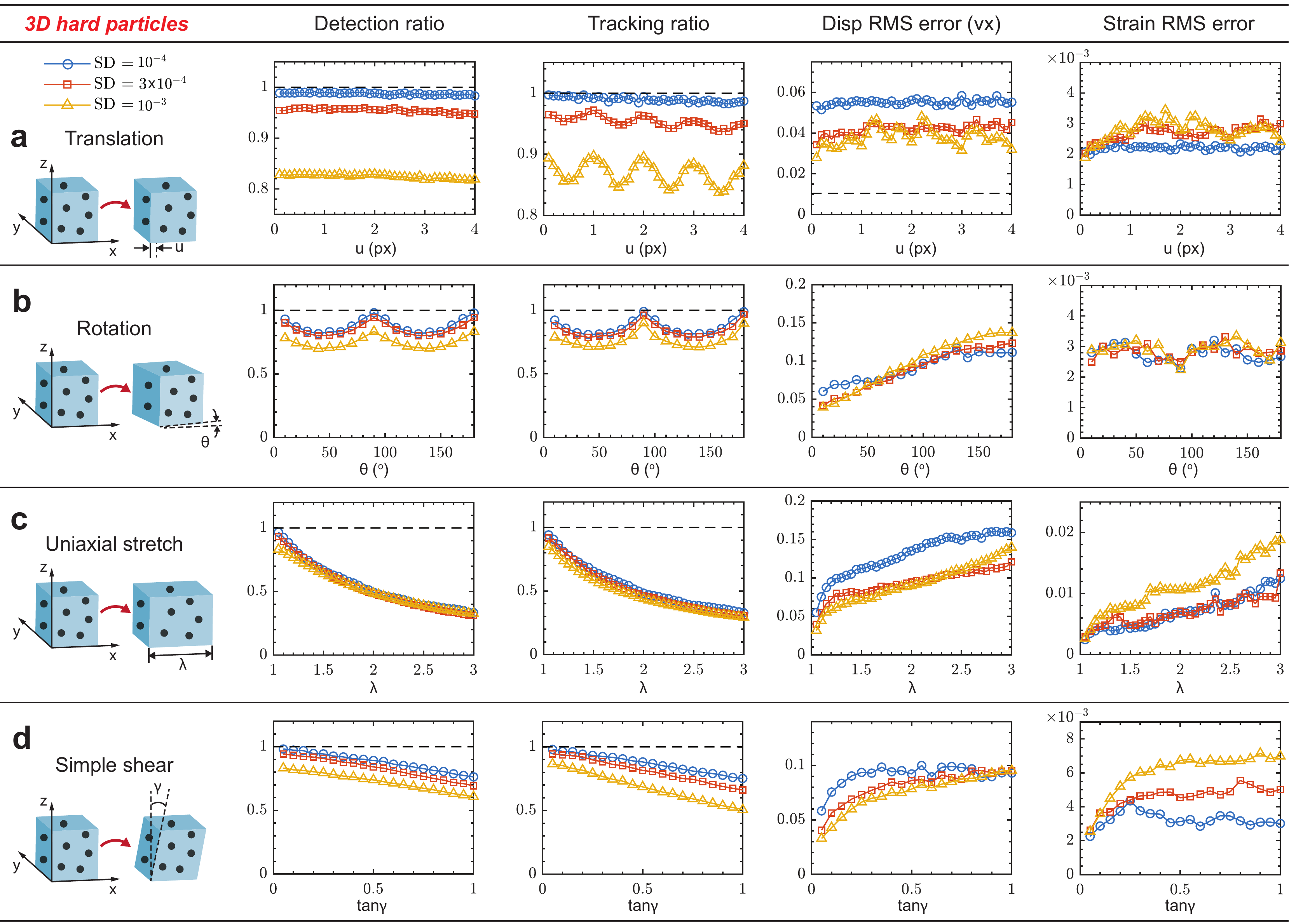}
\caption{The ratio of detected to total ``hard" particles, the overall ratio of successfully tracked particles to total detected particles, and the displacement and strain RMS errors as a function of increasing applied motion, i.e., image step, for synthetic 3D homogeneous deformations for the proposed ALT-SCRIPT method. (a)\,Unidirectional translations in the x-direction from 0 to 4 pixels.
(b)\,Rigid body rotation about the $z$-axis with rotation angles from $0^{\circ}$ to $180^{\circ}$. (c)\,Uniaxial stretch in the $x$-direction with stretch ratios from 1 to 3. (d)\,Simple $xz$-shear with shear angles from $0^{\circ}$ to $45^{\circ}$.}
\label{fig:si_3d_syn}
\end{figure}

\newpage
\section{2D Synthetic uniform deformation: ``soft'' particles}

\begin{figure}[h!]
\centering
\includegraphics[width = 1 \textwidth]{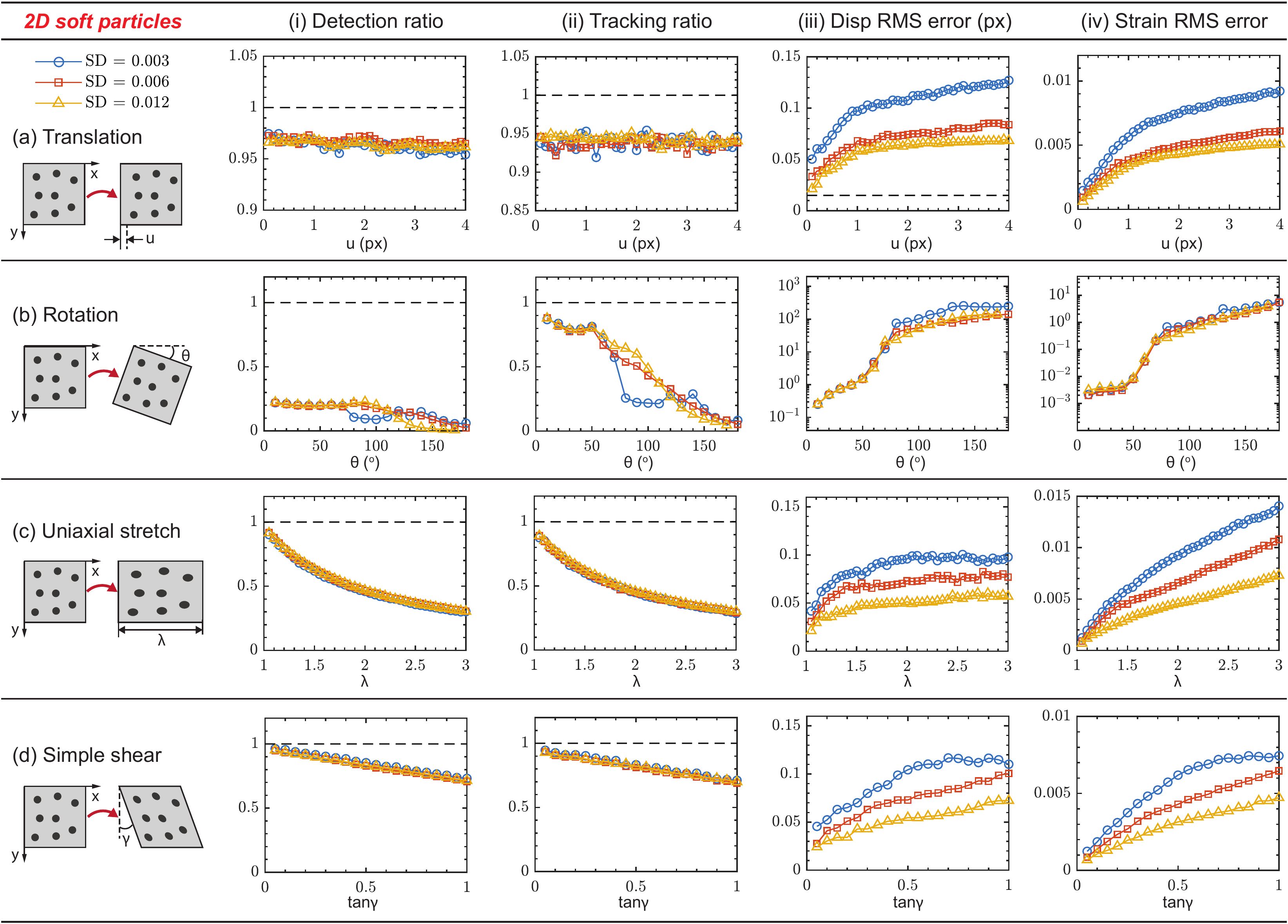}
\caption{The ratio of detected to total ``soft" particles, the overall ratio of successfully tracked particles to total detected particles, and the displacement and strain RMS errors as a function of increasing applied motion, i.e., image step, for synthetic 2D homogeneous deformations for the proposed ALT-SCRIPT method. (a)\,Unidirectional translations in the x-direction from 0 to 4 pixels.
(b)\,Rigid body rotation about the $z$-axis with rotation angles from $0^{\circ}$ to $180^{\circ}$. (c)\,Uniaxial stretch in the $x$-direction with stretch ratios from 1 to 3. (d)\,Simple $xy$-shear with shear angles from $0^{\circ}$ to $45^{\circ}$.}
\label{fig:si_2d_syn_soft}
\end{figure}





